\documentclass{article} 
\usepackage{iclr2026_conference,times}

\usepackage{amsmath,amsfonts,bm}

\def\eqref#1{equation~\ref{#1}}

\def\1{\bm{1}}

\DeclareMathAlphabet{\mathsfit}{\encodingdefault}{\sfdefault}{m}{sl}
\SetMathAlphabet{\mathsfit}{bold}{\encodingdefault}{\sfdefault}{bx}{n}

\usepackage{hyperref}
\usepackage{url}

\usepackage[utf8]{inputenc} 
\usepackage[T1]{fontenc}    
\usepackage{booktabs}       
\usepackage{amsfonts}       
\usepackage{nicefrac}       
\usepackage{microtype}      
\usepackage{xcolor}         
\usepackage{graphicx}
\usepackage{multirow}
\usepackage{graphicx}
\usepackage{xparse}
\usepackage{makecell}
\usepackage{booktabs}
\usepackage{amsmath}
\usepackage{overpic}
\usepackage{xspace}
\usepackage[dvipsnames]{xcolor}

\usepackage[most]{tcolorbox}
\usepackage{caption}
\usepackage{lipsum}

\usepackage{inconsolata}

\usepackage{graphicx}
\usepackage{multirow}
\usepackage{booktabs}
\usepackage[table,xcdraw]{xcolor}
\usepackage{tcolorbox}
\tcbuselibrary{breakable, skins}
\usepackage{xspace}

\usepackage{hyperref}
\usepackage{url}
\usepackage{enumitem}

\tcbset{
  promptstyle/.style={
    enhanced,
    colback=gray!10,
    colframe=black,
    boxrule=0.8pt,
    arc=4mm,
    boxsep=4pt,
    left=4pt,
    right=4pt,
    top=4pt,
    bottom=4pt,
    width=\linewidth,
    fontupper=\ttfamily
  }
}

\newcommand{\et}[2]{${#1}^{\pm{#2}}$}
\newcommand{\name}{{Text2Interact}\xspace}
\newcommand{\namesyn}{{InterCompose}\xspace}
\newcommand{\namegen}{{InterActor}\xspace}

\title{\name: High-Fidelity and Diverse Two-Person Interaction Generation from Text}

\author{Qingxuan Wu$^1$, Zhiyang Dou$^{1,2}$, Chuan Guo$^3$, Yiming Huang$^1$, Qiao Feng$^1$, \\
  \textbf{Bing Zhou$^3$, Jian Wang$^3$, Lingjie Liu$^1$} \\
  $^1$University of Pennsylvania, $^2$The University of Hong Kong, $^3$Snap Inc.
}

\iclrfinalcopy 
\begin{document}

\maketitle

\begin{abstract}
Modeling human–human interactions from text remains challenging because it requires not only realistic individual dynamics but also precise, text-consistent spatiotemporal coupling between agents. Currently, progress is hindered by 1) \textit{limited two-person training data}, inadequate to capture the diverse intricacies of two-person interactions; and 2) insufficiently fine-grained text-to-interaction modeling, where language conditioning collapses rich, structured prompts into a single sentence embedding. To address these limitations, we propose our \name framework designed to generate realistic, text-aligned human–human interactions through a scalable high-fidelity interaction data synthesizer and an effective spatiotemporal coordination pipeline. First, we present \namesyn, a scalable synthesis-by-composition pipeline that aligns LLM-generated interaction descriptions with strong single-person motion priors. Given a prompt and a motion for an agent, \namesyn retrieves candidate single-person motions, trains a conditional reaction generator for another agent, and uses a neural motion evaluator to filter weak or misaligned samples—expanding interaction coverage without extra capture. Second, we propose \namegen, a text-to-interaction model with word-level conditioning that preserves token-level cues (initiation, response, contact ordering) and an adaptive interaction loss that emphasizes contextually relevant inter-person joint pairs, improving coupling and physical plausibility for fine-grained interaction modeling. Extensive experiments show consistent gains in motion diversity, fidelity, and generalization, including out-of-distribution scenarios and user studies. We will release code and models to facilitate reproducibility.
\end{abstract}

\section{Introduction}

Modeling realistic and controllable two-person interactions from natural language remains a central problem in human motion generation with broad implications for animation, extended reality, and embodied AI. Compared with single-person motion, interactions require (i) plausible individual dynamics, (ii) precise spatiotemporal coupling between agents, and (iii) semantic consistency with text prompts that often specify initiation, response, and contact phases. Despite recent progress in single-person synthesis~\citep{tevet2022human, zhou2024emdm, cong2024laserhuman, xu2025mospa, guo2024momask}, current approaches struggle to extend to diverse two-person scenarios for two principal reasons. \emph{Data coverage is insufficient:} two-person corpora are markedly smaller than single-person counterparts (e.g., \textsc{InterHuman}~\citep{liang2024intergen}: $<\!8$k sequences vs.\ \textsc{HumanML3D}~\citep{guo2022generating}: $>\!14$k), which limits the support of coordination patterns and degrades out-of-distribution generalization. \emph{Language conditioning is coarse:} while interaction captions are long and structurally informative (median $21$ tokens in \textsc{InterHuman} vs.\ $7$ in \textsc{HumanML3D}), prior work~\citep{liang2024intergen, tanaka2023role, javed2024intermask} typically compresses the prompt into a single sentence embedding holistically, discarding token-level spatial–temporal cues essential for text-to-interaction alignment. 

In this paper, we argue that in text-to-interaction generation, \emph{coverage can be obtained by scalable motion composition} and \emph{faithfulness can be achieved by fine-grained language supervision and coupling-aware objectives}. To this end, we present \name, a framework that generates realistic, text-aligned human–human interactions via a scalable high-fidelity data synthesizer and a streamlined spatiotemporal coordination pipeline.

First, we introduce \namesyn, our scalable data synthesizer, based on the view that many interactions can be \emph{composed} from single-person motion primitives when composition is text-grounded and constrained by inter-agent geometry and timing. Given a prompt $x=(x_1,\ldots,x_T)$ and an observed (or synthesized) motion for agent~A, we learn a conditional generator for agent~B that enforces semantic alignment with $x$ and spatiotemporal consistency with A. Our scalable synthesis-by-composition framework aligns LLM-generated interaction descriptions with strong single-person motion priors to produce diverse two-person sequences beyond existing distributions. We pair rich interaction texts with concise single-person summaries via an LLM~\citep{liu2024deepseek}, retrieve candidate single-person motions from a pretrained generator~\citep{guo2024momask}, and train a conditional \emph{reaction} model to produce B given A and $x$. A neural motion evaluator filters weak or misaligned samples, yielding synthetic data that broadens interaction support without additional capture.

\begin{figure}
\vspace{-5mm}
\centering

  \begin{overpic}[width=\linewidth]{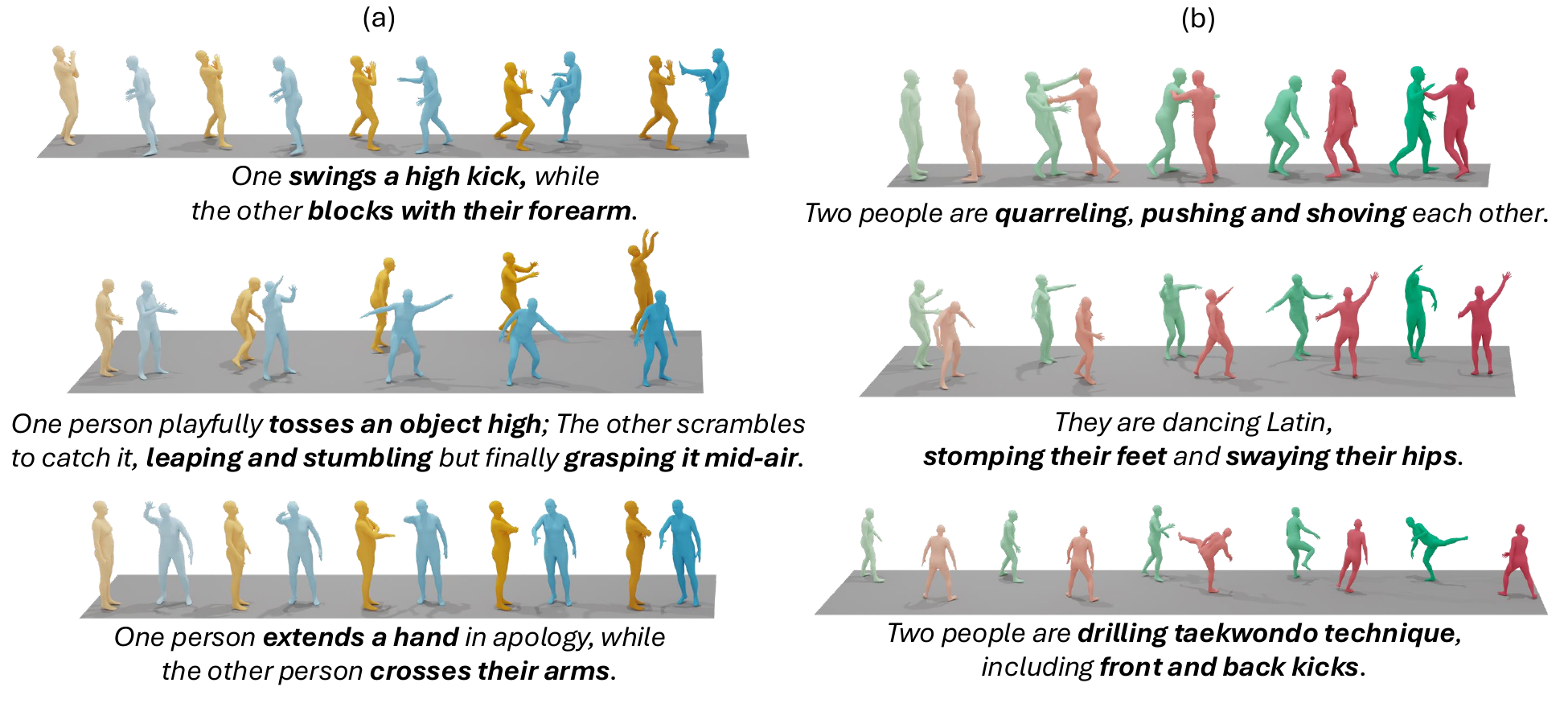}
  \end{overpic}
\vspace{-5mm}
\caption{(a) Our generative two-person motion composition framework, \namesyn, synthesizes plausible and diverse interactions from generated textual descriptions and a single-person motion condition (yellow). (b) Our interaction generation framework \namegen generates high-quality and plausible interactions faithful to text. A deeper color indicates a later time.}
\vspace{-3mm}
\label{fig:fig_teaser}
\end{figure}

Second, we present \namegen, our text-to-interaction generator that excels in language-interaction consistency and spatial-temporal coordination. We achieve fine-grained language-to-interaction modeling by employing \emph{word-level conditioning} rather than sentence-level conditioning~\citep{liang2024intergen, javed2024intermask, ruiz2024in2in} in \namegen, allowing motion tokens to attend to all textual tokens and thus preserve cues about initiation, response, and contact ordering. To further effectively capture the detailed body interactions and facilitate spatial-temporal consistency, we introduce an \emph{adaptive interaction loss} that weights inter-person joint pairs according to context-dependent proximity and relevance (e.g., hands in handshakes, forearms in sparring), in contrast to diffusion-based baselines that treat all pairs uniformly~\citep{liang2024intergen, ruiz2024in2in}.

Extensive experiments demonstrate state-of-the-art performance of our method in the standard benchmark~\citep{liang2024intergen} of two-person motion generation, showing that \namegen outperforms prior art in terms of motion fidelity, faithfulness to text, and generalizability. Moreover, our ablation studies validate the effectiveness of each component in the proposed framework, especially in scenarios where real interaction data is sparse. 

To faithfully and comprehensively measure the effectiveness of the \namesyn framework, we conduct a user preference study that evaluates the improvement on generation quality and text-alignment from in-the-wild text prompts when \namegen is fine-tuned on synthetic data. This measures improvements in generalizability that are undetected by quantitative metrics calculated from the InterGen-trained~\citep{liang2024intergen} evaluator embeddings.

In summary, our contribution is threefold:
\begin{itemize}[leftmargin=*]
    \vspace{-2mm}
    \item A scalable synthesis-and-filtering strategy (\namesyn) that constructs high-quality, diverse two-person interactions from LLM text priors and single-person motion priors.
    \vspace{-2mm}
    \item A word-level attention conditioning module (\namegen) with an adaptive interaction loss for semantically faithful and spatiotemporally coherent two-person generation.
    \vspace{-2mm}
    \item State-of-the-art results on standard benchmarks and superior performance in challenging out-of-distribution settings via a broad user study.
\end{itemize}

\section{Related Works}

\subsection{Text-to-Human Motion Generation}
\vspace{-0.5mm}

Text-to-motion generation aims to synthesize human motion sequences from natural language descriptions~\citep{fan2024freemotion, tanke2023social, jeong2024multi, jiang2023motiongpt, guo2022generating, zhang2023generating, wan2024tlcontrol, lu2024scamo, guo2024momask}. Early methods such as Text2Action~\citep{ahn2018text2action} and Language2Pose~\citep{ahuja2019language2pose} utilized GANs and sequence-to-sequence architectures to map text to motion, laying foundational work in this area. Subsequent approaches leveraged variational autoencoders (VAEs) for probabilistic generation, including Guo et al.~\citep{guo2022generating} and TEMOS~\citep{petrovich2022temos}, which improved motion diversity and fluency. More recent advancements have focused on powerful generative models. Diffusion-based approaches such as MDM~\citep{tevet2022human} and latent diffusion via MLD~\citep{chen2023executing} significantly improved motion realism and sample efficiency. T2M-GPT~\citep{zhang2023generating} employed autoregressive transformers for fine-grained motion synthesis, while MoMask~\citep{guo2024momask} introduced generative masked transformers to enhance fidelity under the autoregressive paradigm. ReMoDiffuse~\citep{zhang2023remodiffuse} further enhanced generation quality by retrieving reference motions from a motion database. Parallel to improving generation quality, increasing attention has been given to controllable text-to-motion generation. Techniques have explored conditioning on spatial trajectories~\citep{shafir2023human, karunratanakul2023guided, wan2024tlcontrol, xie2023omnicontrol} and linguistic constraints~\citep{wan2024tlcontrol, huang2024controllable} to provide more precise control over generated outputs. Additionally, MotionCLIP~\citep{tevet2022motionclip} aligned motion and language embeddings in a shared space, enabling zero-shot text-to-motion generation. Despite stellar results in single-person motion generation, extending them to two-person interactions introduces additional challenges such as modeling inter-agent coordination and handling semantically richer text descriptions. Our work builds on these foundations by proposing a scalable framework that composes diverse and semantically aligned two-person interactions from single-person motion priors and language models.

\subsection{Human-Human Interaction Generation}
\vspace{-0.5mm}
Although some progress has been achieved in multi-human interaction modeling~\citep{fan2024freemotion, tanke2023social, jeong2024multi}, prior works on human interaction modeling have been mostly focused on the two-person interaction problem. A pioneer work, ComMDM~\citep{shafir2023human}, explores two-person motion generation by using a bridge network to compose the outputs of two single-person motion diffusion models~\citep{tevet2022human}. RIG~\citep{tanaka2023role} and InterGen~\citep{liang2024intergen} first trained dedicated networks to directly model two-person interaction. in2IN~\citep{ruiz2024in2in} explores the simultaneous use of individual and interaction descriptions to enhance textual alignment and generation quality. MoMat-MoGen~\citep{cai2024digital} proposes to enhance generation quality by retrieving from a motion database and a generative framework that models interactive behaviors between agents, considering personality, motivations, and interpersonal relationships. InterMask~\citep{javed2024intermask} utilizes the generative masked transformer architecture and spatial-temporal attention to enhance generation quality and text-motion alignment. TIMotion~\citep{wang2024temporal}, a contemporaneous work, proposes to model the human interaction sequence in a causal sequence, leveraging the temporal and causal properties of human motions. Although these methods have achieved impressive results, there remains significant possibilities of improvement due to their common flaw of limited training corpus and inadequate text modeling granularity. In this paper, we aim to tackle these two key issues with our generative interaction composition framework and fine-grained word-level conditioning module.

\vspace{-2mm}
\section{Method}

\begin{figure}
\vspace{-7mm}
\centering
\begin{overpic}[width=0.95\linewidth]{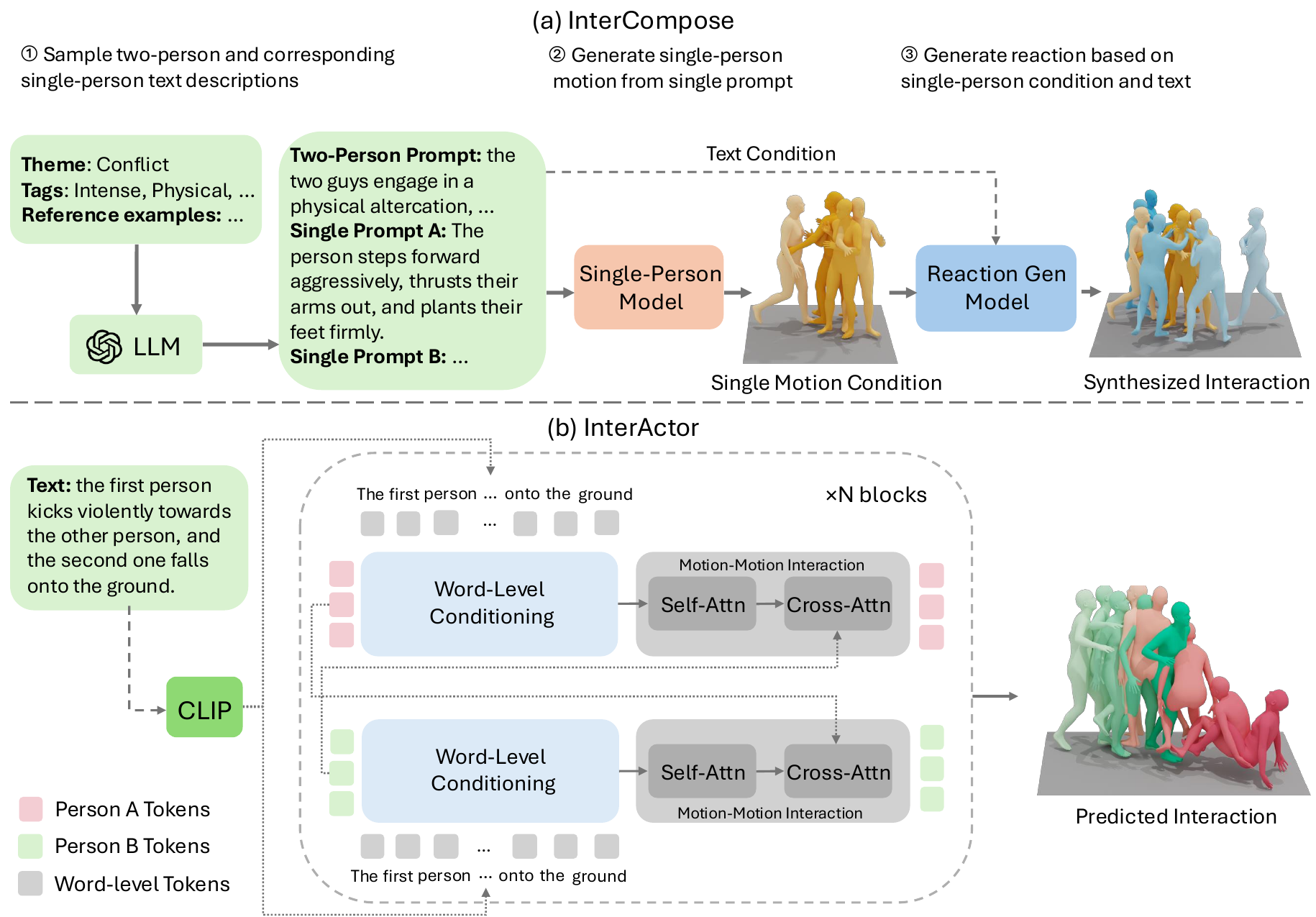}
\end{overpic}

\caption{Overview of the proposed frameworks. (a) \textbf{\namesyn}: sample interaction and single-person descriptions via an LLM, generate a single-person motion from a motion prior~\citep{guo2024momask}, then compose the second agent with a reaction model conditioned on the two-person prompt and the motion prior. (b) \textbf{\namegen}: an $N$-block generator with word-level conditioning and motion–motion interaction. Each block cross-attends motion tokens to CLIP word tokens~\citep{radford2021learning}, followed by self-attention and inter-agent cross-attention to model individual motion and interactions.}

\vspace{-5mm}
\label{fig:fig_pipeline}
\end{figure}

\paragraph{Problem Formulation.} Given a text prompt $c_t$, the task of  \textit{human-human interaction generation from text} involves generating a two-person interaction sequence $\mathbf{X} = [\mathbf{x}_1, \mathbf{x}_2] \in \mathbb{R}^{2 \times T \times N \times 3}$ that is both semantically and spatially coherent and faithful to the original text prompt, where $\mathbf{x}_i$ denotes the i-th person's motion sequence, $T$ is the sequence length in frames and $N$ is the number of joints. Following standard practice in single human and interaction generation \citep{guo2022generating, liang2024intergen, ponce2024in2in}, we use a extended representation formulated as: $\mathbf{x}_{i}^{(t)} = [\mathbf{j}^p_g, \mathbf{j}^v_g, \mathbf{j}^r, \mathbf{c}_f]$, where motion state of the $i$-th person at time $t$, $\mathbf{x}_{i}^{(t)}$, is defined as a collection of global joint positions $\mathbf{j}^p_g \in \mathbb{R}^{3N}$, velocities $\mathbf{j}^v_g \in \mathbb{R}^{3N}$ in the world frame, 6D representation of local rotations $\mathbf{j}^r \in \mathbb{R}^{6N}$ in the root frame, and binary foot-ground contact $\mathbf{c}_f \in \mathbb{R}^4$.

\subsection{\namesyn: Generative Two-person Motion Composition}
\vspace{-0.5mm}
\subsubsection{Data Generation from Single-Person Motion and Language Priors}
\vspace{-0.5mm}

To address the limited diversity of existing two-person motion datasets, we propose a modular pipeline, \namesyn, that synthesizes realistic two-person interactions by sampling coherent two-person and single-person motion descriptions and composing individual motion sequences generated from the descriptions. Specifically, we first use an LLM \citep{liu2024deepseek} to annotate the text descriptions in InterHuman~\citep{liang2024intergen}, classifying them into a discrete space of coarse-grained \emph{themes} (e.g. greeting, dancing, conflict) and fine-grained \emph{tags} (e.g. excited, synchronized, disarm) that further describes the interaction. By systematically combining plausible theme–tag combinations, we can generate interaction descriptions that remain stylistically consistent with InterHuman but span a broader range of behaviors by sampling from the LLM in the joint theme-tag space: $c_t \sim \mathcal{T}_{\text{LLM}}(\text{theme}, \text{tags})$.  Then, given a generated interaction text $c_t$, we decompose it into two role-specific sub-descriptions $(c_t^1, c_t^2)$ using an additional LLM prompt. Each $c_t^i$ describes the motion of person $i$ independent of the other, while taking the context information into account. Please refer to Fig.~\ref{fig:fig_pipeline}~(a) for an illustration. We use $(c_t^1, c_t^2)$ to generate corresponding single-person motions $(\mathbf{x}_1, \mathbf{x}_2)$ via a pre-trained single-person text-to-motion generator MoMask~\citep{guo2024momask} trained on single-person motion datasets~\citep{guo2022generating}, enabling it to generate motions beyond the single-person motion distribution of InterHuman~\citep{liang2024intergen}.

To model dependencies between the interactants, we train a conditional diffusion model $\mathcal{D}_\theta$ that synthesizes the second agent’s motion $\mathbf{x}_2$ given the first agent’s motion $\mathbf{x}_1$ and the shared interaction description $c_t$. Formally, we model the conditional distribution $p_\theta(\mathbf{x}_2 \mid \mathbf{x}_1, c_t)$ using a denoising diffusion probabilistic model (DDPM) with an 8-layer Transformer architecture. At training time, $\mathcal{D}_\theta$ aims to recover an interaction sequence $(\mathbf{x}_1, \mathbf{x}_2)$ sampled from InterHuman~\citep{liang2024intergen} from one ground-truth and one noised interactant $(\mathbf{x}_1, \mathbf{x}_2')$. During inference, we sample $\mathbf{x}_1$ using MoMask then generate $\mathbf{x}_2$ using $\mathcal{D}_\theta$ conditioned on $\mathbf{x}_1$, producing a complete interaction $(\mathbf{x}_1, \mathbf{x}_2)$ that is semantically aligned with $c_t$ and physically coordinated. 

This compositional approach significantly enlarges the diversity of two-person interactions compared to existing datasets, as it decouples single-person motion priors and recombines them under guided conditions. Unlike direct generation approaches, which must learn joint coordination from sparse data, our formulation leverages both rich single-person priors and role-specific semantics to scaffold plausible and varied interactions from structured textual prompts. In addition, the inference-based nature of our data composition process allows it to be extremely scalable and cost-efficient compared to the traditional MoCap-based data collection process.
\vspace{-0.5mm}

\subsubsection{High-Quality and Diverse Data Filtering.} 
\vspace{-0.5mm}

To ensure the quality and diversity of the synthesized two-person motions, we propose a two-stage filtering pipeline that considers text-motion alignment and distributional regularization. We first train a contrastive encoder using the  InterHuman~\citep{liang2024intergen} two-person interaction dataset to project both text and motion into a shared embedding space. Specifically, we freeze a pretrained text encoder (CLIP~\citep{radford2021learning}) with a trainable Transformer~\citep{vaswani2017attention} feature extractor head $f_{head}$, and learn a motion encoder $f_\phi$ based on the Transformer architecture. The training objective is a symmetric cross-entropy (CE) loss over cosine similarities between normalized embeddings. A held-out subset of the InterHuman dataset is reserved to provide a reference embedding distribution for diversity filtering.

After training, we apply the encoder to the synthetic dataset $\mathcal{D}_{\text{syn}} = \{(\mathbf{x}, c_t)\}$ and compute the cosine similarity between each motion and its paired text. We discard samples with similarity scores below a threshold $\delta = 0.58$, empirically chosen based on performance on a validation split. This step eliminates low-quality or semantically misaligned samples.

To further enforce motion diversity and promote high-quality samples that are underrepresented in the original two-person dataset, we perform a distributional filtering step using the two-person motion embeddings from the held-out InterHuman subset $\mathcal{E}_{\text{real}} = \{f_\phi(\mathbf{x}_r)\}$ as reference. For each synthetic motion embedding $f_\phi(\mathbf{x})$, we compute its Euclidean distance to the $k$ nearest neighbors in $\mathcal{E}_{\text{real}}$, and retain only those whose average distance falls within a predefined annulus: $r_{\text{min}} \leq d(f_\phi(\mathbf{x}), \mathcal{E}_{\text{real}}) \leq r_{\text{max}}$. This preserves synthesized motions that are novel (outside the inner radius $r_{\text{min}}$) but not far from the real data distribution (inside the outer radius $r_{\text{max}}$).

This dual-stage filtering framework ensures that the final synthetic dataset exhibits both semantic fidelity and distributional diversity. Detailed analysis of the effects of $\delta$, $r_{\text{min}}$, and $r_{\text{max}}$ is in Sec.~\ref{sec:filtering_ablation}.

\subsection{\namegen: Fine-grained Interaction Modeling}

\subsubsection{Word-level Attention Modeling of Language and Interaction Dynamics}

Having diversified our training distribution with synthetic data, we now address the issue of insufficient granularity in two-person text semantics modeling. To tackle the issue and improve semantic alignment between natural language and generated motion, we design a cross-attention-based word-level text-motion conditioning architecture that injects fine-grained text information throughout the generation process. Unlike prior methods that inject a sentence-level embedding into motion tokens via AdaLN~\citep{liang2024intergen,javed2024intermask,ponce2024in2in} or sentence-level cross-attention~\citep{tanaka2023role}, our architecture allows each motion token to dynamically attend to individual word-level tokens, preserving the nuanced motion semantics and spatial-temporal alignment cues in semantic-rich interaction prompts.

Formally, given a tokenized interaction description $c_t = \{w_1, \dots, w_L\}$, we extract word-level embeddings $\mathbf{T} = \{\mathbf{t}^{(1)}, \dots, \mathbf{t}^{(L)}\}$ using a frozen CLIP text encoder.

The architecture is composed of alternating processing modules, each consisting of two types: 1) \textbf{Word-level Conditioning Module $\mathcal{M}_w$:} A Transformer block with cross-attention between a single agent’s motion tokens $\mathbf{x}_i$ and the full text embedding sequence $\mathbf{T}$, enabling each motion token to focus on semantically relevant parts of the prompt. This block preserves temporal resolution and injects lexical cues aligned with event structure. 2) \textbf{Motion-Motion Interaction Module $\mathcal{M}_m$:} A two-stage module where motion tokens $\mathbf{x}_i$ first perform self-attention over their own sequence (intra-agent context), followed by cross-attention over the other agent’s motion tokens $\mathbf{x}_{j}$ ($j \neq i$), which models inter-agent physical and temporal dependencies such as push-pull or synchronization. Please see Fig.~\ref{fig:fig_pipeline}~(b) for an illustration.

Each update step consists of a word-level conditioning module followed by a motion-motion interaction module; these two modules together form a full block that is applied in an alternating fashion: first to one agent, conditioning on the text and the other agent’s motion, and then to the other agent in the next step. Leveraging the symmetry of two-person interactions, the blocks $\mathcal{B}_w$ and $\mathcal{B}_m$ are shared across agents, ensuring architectural symmetry and parameter efficiency. The alternating structure allows each agent to respond adaptively to both the linguistic description and the dynamic behavior of their partner, while preserving causal and temporal coherence.

Overall, the network design enables high-fidelity generation that is both semantically grounded and interaction-aware, allowing nuanced conditioning through the word-level representation and fostering motion patterns that are faithful to the described scenario.

\subsubsection{Adaptive Interaction Supervision}

We use the standard velocity loss $\mathcal{L}_{\text{vel}}$, foot contact loss $\mathcal{L}_{\text{foot}}$, bone-length loss $\mathcal{L}_{\text{BL}}$, and relative orientation loss $\mathcal{L}_{\text{RO}}$. For these objective functions, refer to InterGen~\citep{liang2024intergen} for details. 

In addition to the above objective functions, we designed a new objective $\mathcal{L}_{\text{AdaInteract}}$ to enhance the generation of plausible interaction semantics, a crucial element of text-to-interaction generation. Motivated by the insight that joint pairs that are closer to each other carry more importance in the interaction semantics, we propose a novel adaptive interaction loss that supervises the pairwise distances between human-human joint pairs with an adaptive weighting:
\begin{equation}
    \mathcal{L}_{\text{AdaInteract}} = \sum_{i=1}^N\sum_{j=1}^N\frac{1}{d_{ij} + \epsilon}\|d_{ij} - \hat d_{ij}\|_2
\end{equation}
Where $d_{ij}$, $\hat d_{ij}$ are the ground-truth and predicted distances between the joints $i$ and $j$ respectively, and $\epsilon=0.1$ is an empirically set constant. By putting more emphasis on spatially proximate inter-agent joint pairs, our adaptive interaction objective function provides strong guidance for the model to adhere to the interaction semantics.

\section{Experiments}

\subsection{Experimental Setup}

\textbf{Dataset.} We use the InterHuman~\citep{liang2024intergen} dataset for training and evaluating our model. InterHuman contains 6,022 two-person interacting motions and 3 textural descriptions per motion in the training split, and 1,177 two-person interacting motions in the test split. Additionally, a synthesized dataset of 25,000 text-motion pairs before filtering and 1,200 text-motion pairs after filtering is used for fine-tuning. All models are first trained on the InterHuman training split. For fine-tuning, the model is fine-tuned on the InterHuman training split augmented by the filtered synthetic dataset. All metrics are calculated using the InterHuman test split. 

\textbf{Metrics.} Following standard practice in human-human interaction generation~\citep{liang2024intergen,ruiz2024in2in,javed2024intermask,cai2024digital}, we use the R-Precision (Top-1, 2, 3), Frechet Inception Distance (FID), Multimodal Distance (MM Dist), Diversity, and Multimodality (MModality) for evaluation our models. Please refer to InterGen~\citep{liang2024intergen} for the detailed definition of these metrics.

\textbf{Implementation Details.} Our model consists of 12 attention blocks and 12 word-level conditioning blocks, positioned in an interleaved manner. We utilize a frozen CLIP-ViT-L/14~\citep{radford2021learning} model for extracting as the text encoder. We set the number of diffusion~\citep{ho2020denoising} steps to 1,000 and use a cosine noise schedule~\citep{nichol2021improved}. The model is trained with 8 NVIDIA A100 GPUs for 200,000 steps, with a 5e-5 learning rate and a batch size of 16 with the AdamW~\citep{loshchilov2017decoupled} optimizer, cosine learning rate scheduling, and 1000-step warm-up. During sampling, we use the DDIM~\citep{song2020denoising} sampling with 50 timesteps, with a classifier-free guidance~\citep{ho2022classifier} weight of 3.5.

\begin{table*}[t]
\vspace{-6mm}
    \centering
    \makebox[\textwidth]{ 
    \scalebox{0.7}{
    \begin{tabular}{ l c c c c c c c}
    \toprule
\multirow{2}{*}{Method}  & \multicolumn{3}{c}{R Precision$\uparrow$} & \multirow{2}{*}{FID$\downarrow$} & \multirow{2}{*}{MM Dist$\downarrow$} & \multirow{2}{*}{Diversity$\rightarrow$} & \multirow{2}{*}{MModality$\uparrow$} \\
    \cmidrule{2-4}
    ~ & Top 1 & Top 2 & Top 3 \\
    \midrule
         Ground Truth & \et{0.452}{.008} & \et{0.610}{.009} & \et{0.701}{.008} & \et{0.273}{.007} & \et{3.755}{.008} & \et{7.948}{.064} & - \\
    \midrule
         T2M~\citep{guo2022generating} & \et{0.238}{.012} & \et{0.325}{.010} & \et{0.464}{.014} & \et{13.769}{.072} & \et{5.731}{.013} & \et{7.046}{.022} & \et{1.387}{.076} \\
         MDM~\citep{tevet2022human} & \et{0.153}{.012} & \et{0.260}{.009} & \et{0.339}{.012} & \et{9.167}{.056} & \et{7.125}{.018} & \et{7.602}{.045} & \et{\textbf{2.350}}{.080} \\
         ComMDM~\citep{shafir2023human} & \et{0.223}{.009} & \et{0.334}{.008} & \et{0.466}{.010} & \et{7.069}{.054} & \et{6.212}{.021} & \et{7.244}{.038} & \et{1.822}{.052} \\
         RIG~\citep{tanaka2023role} & \et{0.285}{.010} & \et{0.409}{.014} & \et{0.521}{.013} & \et{6.775}{.069} & \et{5.876}{.002} & \et{{7.311}}{.043} & \et{2.096}{.065} \\
         InterGen~\citep{liang2024intergen} & \et{0.371}{.010} & \et{0.515}{.012} & \et{0.624}{.010} & \et{5.918}{.079} & \et{5.108}{.014} & \et{7.387}{.029} & \et{ {2.141}}{.063} \\
         MoMat-MoGen~\citep{cai2024digital} & \et{0.449}{.004} & \et{0.591}{.003} & \et{0.666}{.004} & \et{5.674}{.085} & \et{3.790}{.001} & \et{8.021}{.350} & \et{1.295}{.023} \\
         in2IN~\citep{ruiz2024in2in} & \et{0.425}{.008} & \et{0.576}{.008} & \et{0.662}{.009} & \et{5.535}{.120} & \et{3.803}{.002} & \et{{7.953}}{.047} & \et{1.215}{.023} \\
         InterMask~\citep{javed2024intermask} & \et{0.449}{.004} & \et{0.599}{.005} & \et{0.683}{.004} & \et{\textbf{5.154}}{.061} & \et{3.790}{.002} & \et{\textbf{7.944}}{.033} & \et{1.737}{.020} \\
            \midrule
        
        Ours  & \et{\textbf{0.483}}{.005} & \et{\textbf{0.638}}{.005} & \et{\textbf{0.717}}{.005} & \et{{5.191}}{.055} & \et{\textbf{3.778}}{.001} & \et{7.900}{.030} & \et{1.051}{.031} \\

    \bottomrule
    \end{tabular}
    }
    }
    \caption{Performance on the InterHuman~\citep{liang2024intergen} test sets. $\pm$ indicates a 95\% confidence interval and $\rightarrow$ means the closer to ground truth the better. Boldface indicates the best result.}
    \label{tab:quantitative_eval}
    \vspace{-4mm}
\end{table*}
\subsection{Comparison with the State-of-the-arts}

\paragraph{Quantitative Comparison.} Tab.~\ref{tab:quantitative_eval} contains the quantitative comparison between \namegen and state-of-the-art methods. Each experiment is repeated 20 times, after which the mean and 95\% confidence interval of each metric is recorded. \namegen achieves state-of-the-art results on all three R-precision metrics, surpassing the previous state-of-the-art, InterMask~\citep{javed2024intermask}, by a significant margin, highlighting the effectiveness of the word-level conditioning design choice in text-motion alignment. In terms of motion quality, our \namegen also achieves the best MM Distance and the second-best FID, with a small FID margin (0.037) from the state-of-the-art, InterMask, and surpassing all other prior arts. Notably, our FID is within the 95\% confidence interval of InterMask's FID, highlighting an equal level of generation quality while exhibiting significantly improved R-Precision.

\begin{figure*}[htbp]
\centering
\makebox[\textwidth][c]{
  \begin{overpic}[width=\linewidth]{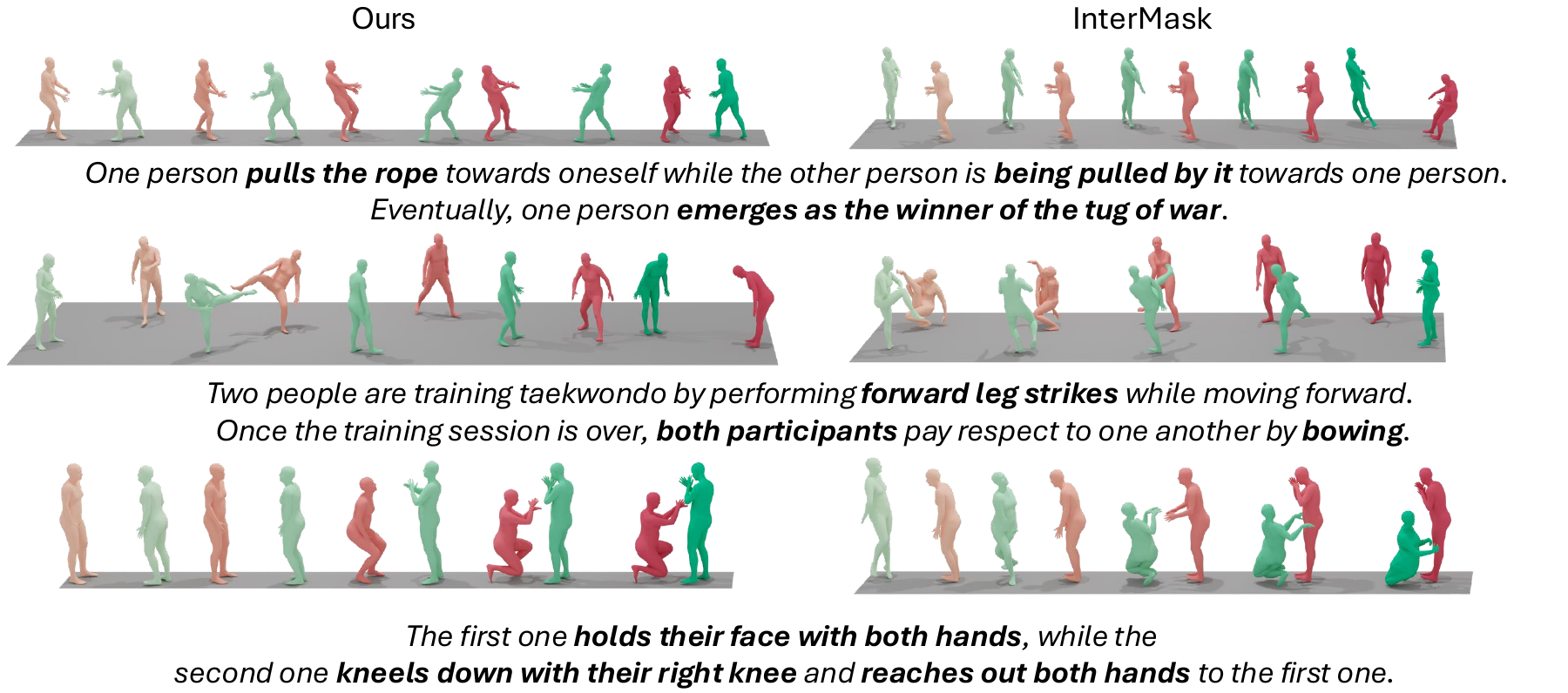}
  \end{overpic}
}
\caption{Qualitative comparisons of interaction generation results from \namegen and InterMask~\citep{javed2024intermask}. Our method produces results with better text-motion alignment and is more robust to implausible poses. A deeper color indicates a later time.}
\label{fig:baseline_compare}
\end{figure*}

\paragraph{Qualitative Comparison.} We provide a qualitative comparison between our method and the current state-of-the-art, InterMask~\citep{javed2024intermask}. As shown in Fig.~\ref{fig:baseline_compare}, our model exhibits stronger adherence to text, higher robustness, and more plausible interaction semantics. 

Specifically, in the first generation result of our method, the agent marked in red successfully pulls the agent marked in green, and the red agent emerges as the winner (frame 5). In contrast, the motion generated by InterMask only exhibits pulling, and does not reflect this final part of the motion. In the second row, the InterMask result exhibits implausible human pose outputs in frames 1 and 2, and does not reflect the final bowing action, while our model generates plausible results faithful to the complete semantic meanings of the text. In the third row, the kneeling human generated by InterMask again exhibits an implausible human pose in frames 2, 3, 4, and 5. These results highlight our \namegen's pose robustness over InterMask, an aspect not adequately measured by the evaluator and the FID metric, while confirming \namegen's lead in text to motion alignment.

\subsection{Further Evaluation}
\begin{table*}[htpb]
\vspace{-2mm}
    \centering
    \makebox[\textwidth]{ 
    \scalebox{0.7}{
    \begin{tabular}{ l c c c c c c c}
    \toprule
\multirow{2}{*}{Method}  & \multicolumn{3}{c}{R Precision$\uparrow$} & \multirow{2}{*}{FID$\downarrow$} & \multirow{2}{*}{MM Dist$\downarrow$} & \multirow{2}{*}{Diversity$\rightarrow$} & \multirow{2}{*}{MModality$\uparrow$} \\
    \cmidrule{2-4}
    ~ & Top 1 & Top 2 & Top 3 \\

    \midrule
        Before Fine-tuning & \et{0.485}{.010} & \et{0.644}{.007} & \et{0.721}{.009} & \et{5.701}{.065} & \et{3.777}{.001} & \et{7.904}{.033} & \et{1.081}{.019} \\
            \midrule
          Finetune ($0.25 < d < 0.6$) & \et{0.485}{.004} & \et{0.641}{.004} & \et{0.717}{.004} & \et{5.981}{.056} & \et{3.778}{.001} & \et{7.946}{.028} & \et{1.080}{.026} \\
       Finetune ($0.3 < d < 0.6$) & \et{0.480}{.007} & \et{0.635}{.004} & \et{0.715}{.004} & \et{5.682}{.100} & \et{3.779}{.002} & \et{7.909}{.030} & \et{1.058}{.030} \\
        Finetune ($0.35 < d < 0.6$) & \et{0.483}{.005} & \et{0.638}{.005} & \et{0.717}{.005} & \et{5.191}{.055} & \et{3.778}{.001} & \et{7.900}{.030} & \et{1.051}{.031} \\
    \bottomrule
    \end{tabular}
    }
    }
    \caption{Quantitative Results of \namegen after fine-tuning on synthetic data generated by \namesyn. $d$ denotes the Euclidean distance between a synthetic data sample point and its closest held-out data point in the embedding space of the neural evaluator.}
    \label{tab:ablation_sampling}
\end{table*}

\textbf{Quantitative Evaluation of Fine-Tuning and Filtering.} 
\label{sec:filtering_ablation} We present the results of fine-tuning the \namegen model on a combined dataset consisting of InterHuman~\citep{liang2024intergen} training split and filtered synthetic data for 50k steps, with a learning rate of 5e-6. As shown in Tab.~\ref{tab:ablation_sampling}, the model exhibits a similar level of text-to-motion matching (R-Precision) after fine-tuning and a significantly improved FID in the best case, highlighting the improvement in generalizability. Notably, the FID exhibits a clear increasing trend when the minimum Euclidean distance $d$ of the filtering process is increased within a reasonable range, confirming the effectiveness of the proposed filtering pipeline in achieving synthetic data quality and diversity at the same time. The results also indicates that the increased dataset diversity by synthetic data improves the model's generalizability.

\begin{figure*}[t]
\centering
\makebox[\textwidth][c]{
  \begin{overpic}[width=\linewidth]{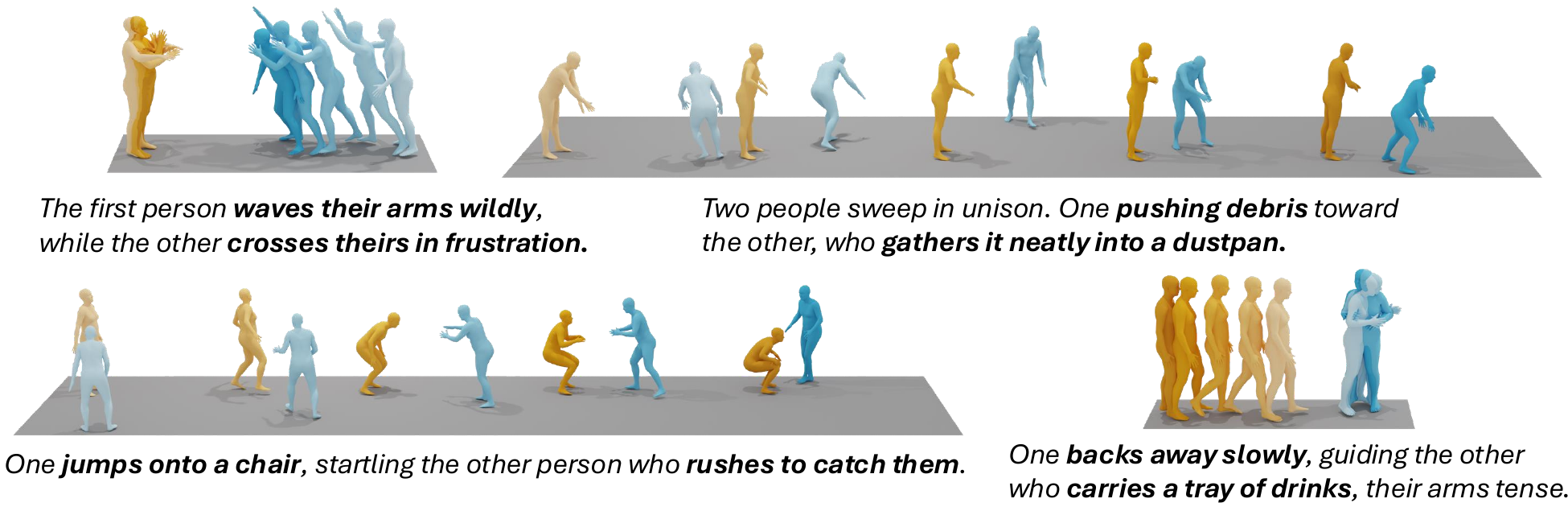}
  \end{overpic}
}
\caption{Qualitative samples of \namesyn. Prompts are synthesized by an LLM~\citep{liu2024deepseek}. The yellow is synthesized by the single-person motion generator, while the blue is generated by the reaction model with the yellow as the condition. A deeper color indicates a later time.}
\label{fig:fig_synthetic}
\end{figure*}

\textbf{Qualitative Visualization of Synthetic Data.} 
In Fig.~\ref{fig:fig_synthetic}, we present exemplar results of our data synthesis pipeline, \namesyn. The agent marked yellow is generated by the single-person motion generator~\citep{guo2024momask} while the agent marked blue is generated by the reaction generation model. As shown in the figure, our pipeline synthesizes high-quality and diverse motions from single-person motion and text descriptions, with close adherence to the text. Moreover, the textual descriptions resemble real-life human-human interaction situations with emotional interactions or inter-person collaboration instilled into the text prompts. 

\begin{figure}[htpb]
\vspace{-1mm}
\centering
\begin{minipage}[t]{0.35\linewidth}
  \centering
  \includegraphics[width=\linewidth]{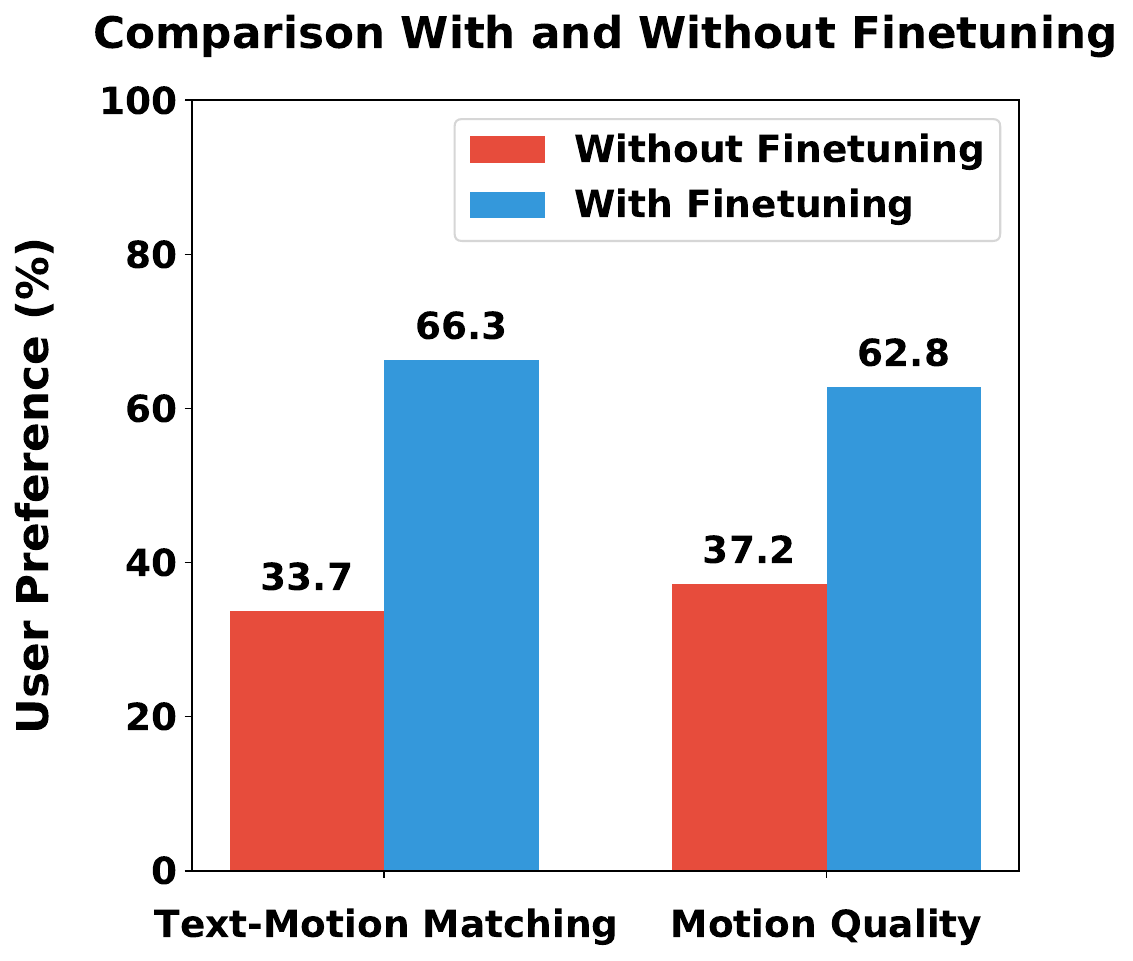}
  \caption{User preference study results of \namegen with and without fine-tuning on synthetic data.}
  \label{fig:finetuning_user_study}
\end{minipage}
\hfill
\begin{minipage}[t]{0.62\linewidth}
  \centering
  \includegraphics[width=\linewidth]{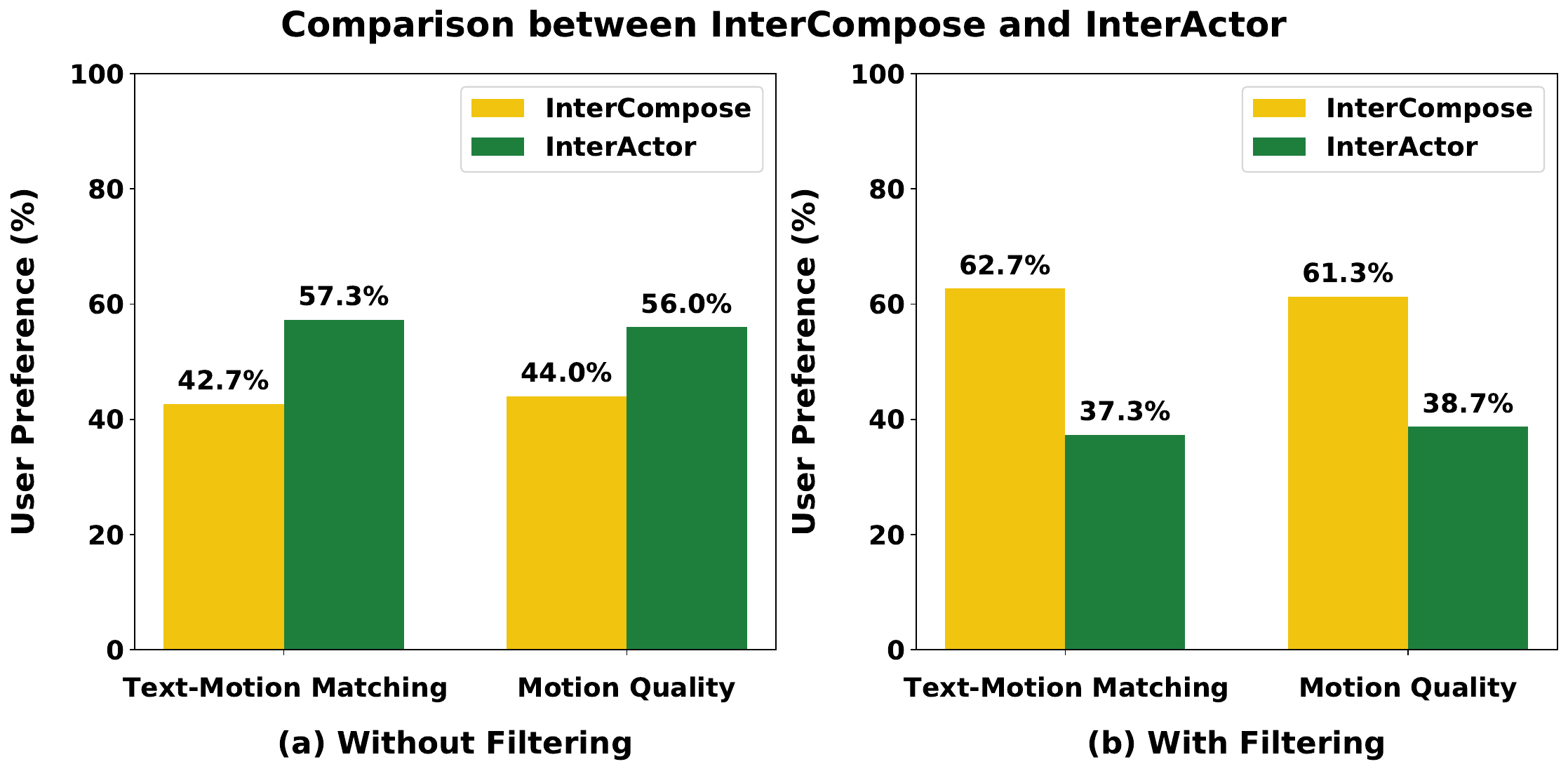}
  \caption{Comparison of motion generation results using \namesyn and \namegen, (a) without filtering, and (b) with filtering. The motion quality and text-motion matching of \namesyn surpass \namegen only after filtering.}
  \label{fig:filtering_user_study}
  \vspace{-3mm}
\end{minipage}

s\end{figure}

\textbf{User Study Results on Fine-Tuning.} 
We present the user preference study results conducted with 51 participants on 10 samples generated with out-of-distribution texts using our data generation pipeline. Fig.~\ref{fig:finetuning_user_study} shows the users' strong preference for the model after fine-tuning, in terms of both motion quality and text-motion matching. This result confirms our model's improved generalizability to out-of-distribution samples after fine-tuning.

\textbf{User Study Results on \namesyn vs \namegen.} Fig.~\ref{fig:filtering_user_study} presents a user preference study between motions synthesized with \namegen and generated by \namegen. Two studies are conducted for \namesyn  (a) without and (b) with distributional filtering, where only high-quality and novel motions are retained after the filtering process. The results show that: (a) \namegen shows stronger consistency in motion quality compared to \namesyn, rendering the former suitable for general text-to-interaction tasks; (b) with the distributional filtering step, motions from \namesyn have higher quality compared to motions generated by \namegen, confirming the quality of the synthesized and filtered motion dataset used for fine-tuning.

\textbf{Ablation Study.} After thoroughly investigating the effectiveness of the data synthesis and fine-tuning pipeline, we now analyze the effectiveness of the proposed sub-components: the word-level conditioning module (WLC), the adaptive interaction loss (AIL), and the synthetic data fine-tuning process (FT). When removing the word-level conditioning module, we replace it with sentence-level condition injection by AdaLN; for ablation of adaptive interaction loss, we replace it with the flat-weight distance map function $\mathcal{L}_{DM}$ proposed by InterGen~\citep{liang2024intergen}; for ablation of fine-tuning, the model was trained only on InterHuman~\citep{liang2024intergen} without the fine-tuning process.
Tab.~\ref{tab:ablation_component} shows significant improvement of our model after adding each proposed component, in terms of text-motion matching (R-Precision), FID, Multimodal Distance, and diversity. 

\begin{table*}[htpb]
    \centering
    \makebox[\textwidth]{ 
    \scalebox{0.7}{
    \begin{tabular}{ l c c c c c c c}
    \toprule
\multirow{2}{*}{Method}  & \multicolumn{3}{c}{R Precision$\uparrow$} & \multirow{2}{*}{FID$\downarrow$} & \multirow{2}{*}{MM Dist$\downarrow$} & \multirow{2}{*}{Diversity$\rightarrow$} & \multirow{2}{*}{MModality$\uparrow$} \\
    \cmidrule{2-4}
    ~ & Top 1 & Top 2 & Top 3 \\
    \midrule

        w.o. All Proposed Components & \et{0.441}{.006} & \et{0.608}{.005} & \et{0.681}{.005} & \et{6.237}{.071} & \et{3.781}{.001} & \et{7.959}{.035} & \et{1.068}{.022} \\

        w.o. AIL, FT & \et{0.484}{.005} & \et{0.632}{.005} & \et{0.710}{.005} & \et{6.192}{.069} & \et{3.779}{.001} & \et{7.853}{.033} & \et{1.081}{.019} \\

        w.o. WLC, FT & \et{0.484}{.005} & \et{0.629}{.005} & \et{0.711}{.005} & \et{5.877}{.061} & \et{3.779}{.001} & \et{7.851}{.034} & \et{0.996}{.027} \\

        w.o. FT & \et{0.485}{.010} & \et{0.644}{.007} & \et{0.721}{.009} & \et{5.701}{.065} & \et{3.777}{.001} & \et{7.904}{.033} & \et{1.046}{.022} \\

        Ours & \et{0.483}{.005} & \et{0.638}{.005} & \et{0.717}{.005} & \et{5.191}{.055} & \et{3.778}{.001} & \et{7.900}{.030} & \et{1.051}{.031} \\

    \bottomrule
    \end{tabular}
    }
    }
    \caption{Ablation Study: Effect of removing one or more of the proposed components: Adaptive Interaction Loss (AIL), Synthetic Data Fine-Tuning (FT), Word-Level Conditioning (WLC).}
    \label{tab:ablation_component}
    \vspace{-3mm}
\end{table*}

\vspace{-0.5mm}
\section{Conclusion}
\vspace{-0.5mm}
In this paper, we presented \name, comprising 1) \namesyn, a novel and effective framework that composes single-person motions into two-person interaction from LLM-generated text descriptions; and 2) \namegen, a high-quality and fine-grained two-person interaction generation framework equipped with word-level conditioning. The effectiveness of \namesyn has been confirmed by an ablation study, user study, qualitative results, and latent visualizations. Utilizing data generated by \namesyn, \namegen achieves a significant FID boost, achieving SoTA-level R-precision and FID, setting a new state-of-the-art for the two-person motion generation task.

\noindent
\textbf{Limitations and Future Work.} While \namegen demonstrates strong motion fidelity and faithfulness, it does not account for physical plausibility during generation, which can result in artifacts such as floating motions and ground penetration. Incorporating physics priors~\citep{peng2018deepmimic, peng2022ase, dou2023c, jiang2023drop, luo2023universal, zhang2024physpt, huang2025modskill} offers a promising avenue for future work. Additionally, although \namesyn provides an effective synthesis framework, extending it to learn motions directly from video~\citep{li2023coordinate, huang2024closely, qiu2023psvt, yuan2022glamr} remains an interesting and relatively unexplored direction.

\newpage

\bibliography{main}
\bibliographystyle{iclr2026_conference}

\clearpage
\appendix
\section*{\textbf{SUPPLEMENTARY MATERIALS}}

\section{\namegen Implementation Details}
\subsection{Word-level Tokenization}
We use the CLIP~\citep{radford2021learning} ViT-L/14 encoder for encoding the text. The text is tokenized by the CLIP tokenizer into word-level tokens for short words and sub-word-level tokens for long words, with \texttt{<SOT>} and \texttt{<EOT>} tokens inserted at the start and end of the text. The maximum number of text tokens is $75$. If the number of tokens after tokenization is longer than $75$, the text tokens are truncated and additional tokens are discarded. 

\subsection{Word-level Conditioning}

\begin{figure*}[htbp]
\centering
\makebox[\textwidth][c]{
  \begin{overpic}[width=0.6\linewidth]{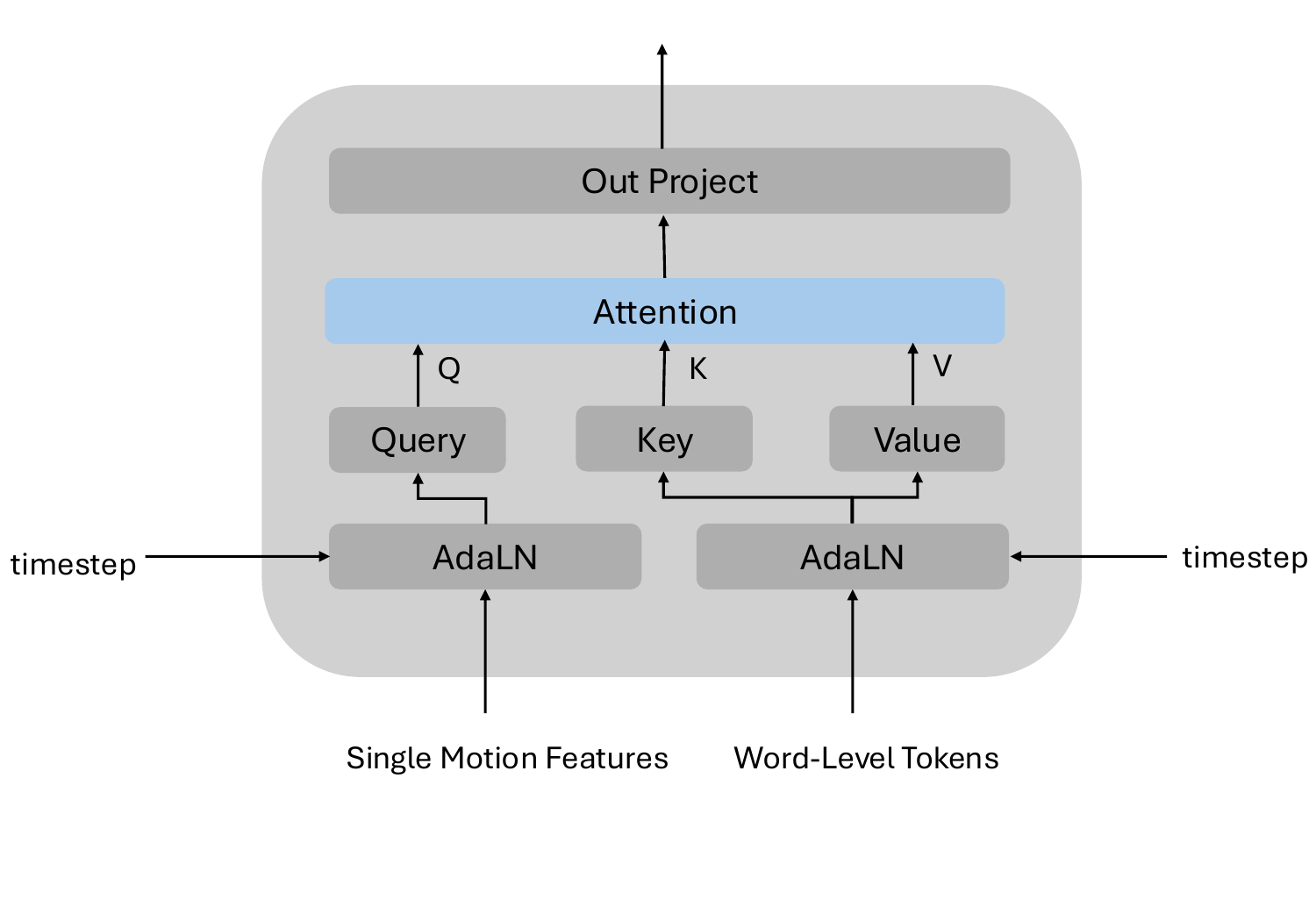}
  \end{overpic}
}
\vspace{-10mm}
\caption{Illustration of the \textbf{Word-Level Conditioning} Block}
\label{fig:fig_appendix_word_level}
\end{figure*}

Fig.~\ref{fig:fig_appendix_word_level} illustrates the details of the proposed Word-Level Conditioning block. One of the interacting agents' motion features (Single Motion Features) is passed through an Adaptive Layer Norm (AdaLN) for the injection of timestep information. The word-level tokens are passed through a separate AdaLN of the same structure but with different parameters. Then, the normalized and modulated features are passed through the linear layers to yield the query, key, and value tensors, where the query comes from the single motion features and the key and value come from the word-level tokens. Then, an attention output embedding is obtained using query $Q$, key $K$, and value $V$ with the attention mechanism~\citep{vaswani2017attention}:
\begin{equation}
\text{Attention}(Q, K, V) = \text{softmax}\left( \frac{QK^\top}{\sqrt{d_k}} \right) V
\end{equation}
Finally, the output of the attention layer is passed through a linear layer for the reprojection and mixing of the attention head outputs.

\subsection{Motion-Motion Interaction}

\begin{figure}[htpb]
\vspace{-1mm}
\centering
\begin{minipage}[t]{0.48\linewidth}
  \centering
  \includegraphics[width=\linewidth]{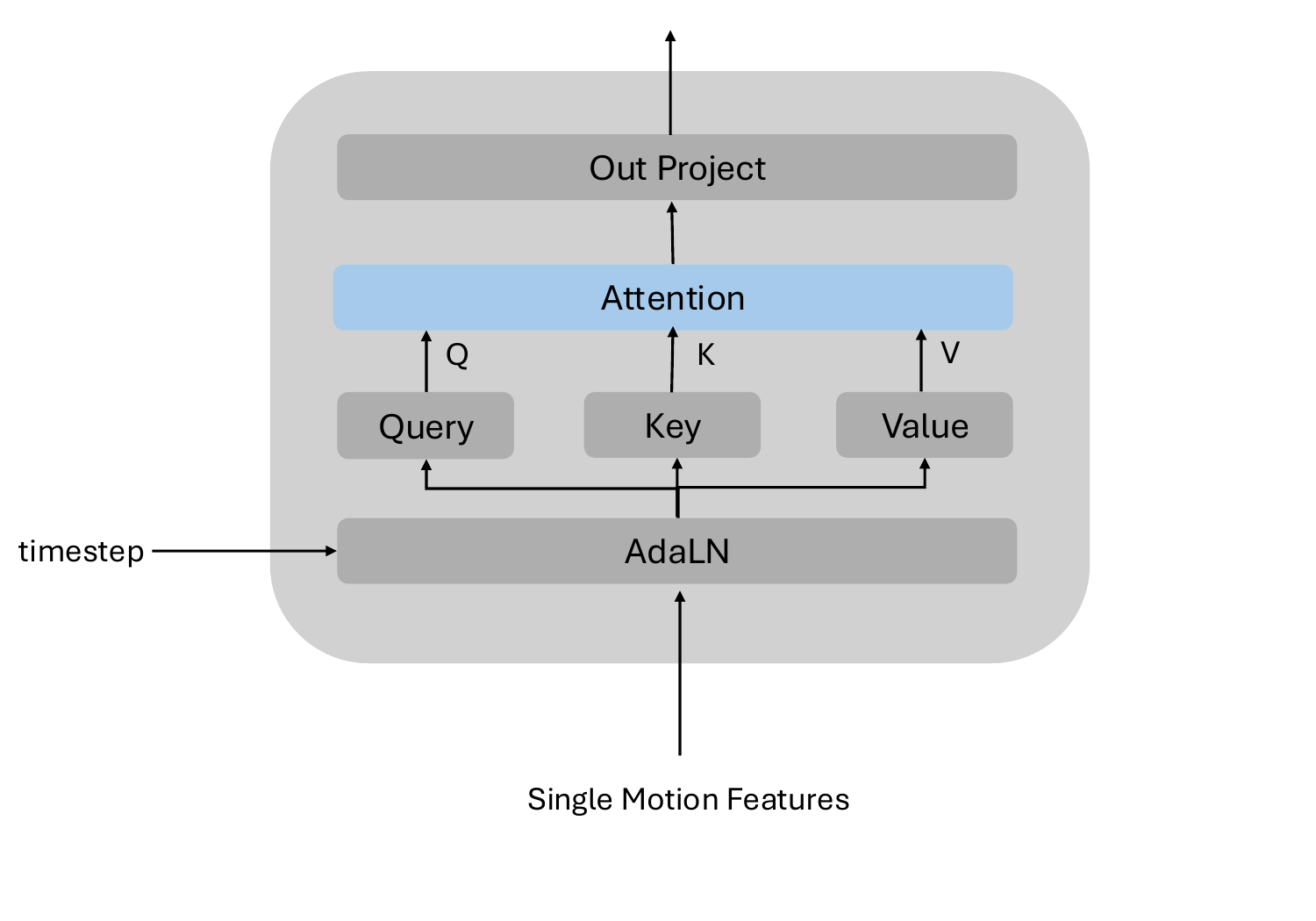}
  \caption{Illustration of the \textbf{Self-Attention} module in the Motion-Motion Interaction Block.}
  \label{fig:fig_appendix_self_attn}
\end{minipage}
\hfill
\begin{minipage}[t]{0.48\linewidth}
  \centering
  \includegraphics[width=\linewidth]{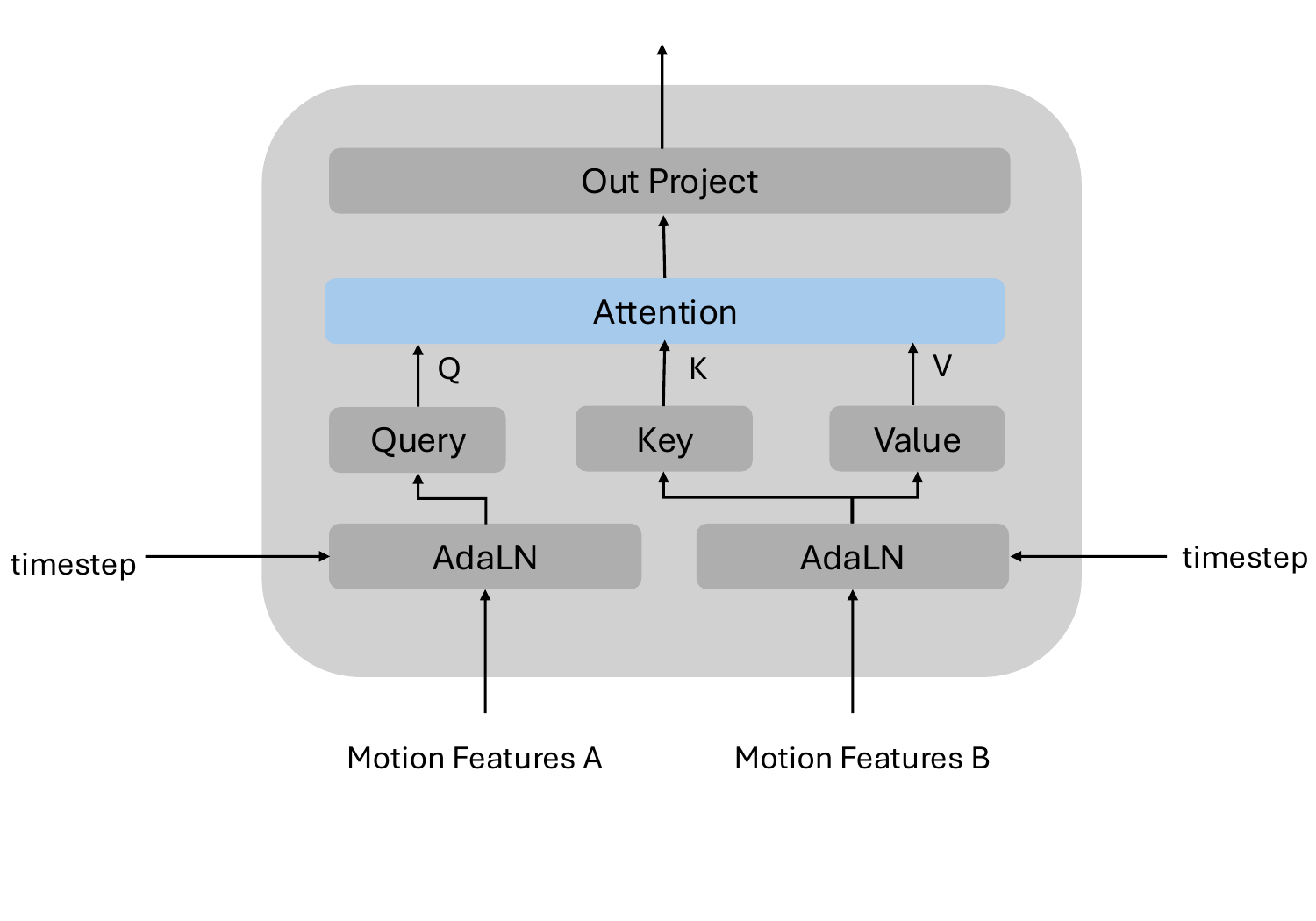}
  \caption{Illustration of the \textbf{Cross-Attention} module in the Motion-Motion Interaction Block.}
  \label{fig:fig_appendix_cross_attn}
  \vspace{-3mm}
\end{minipage}
\end{figure}

\subsubsection{Self-Attention}
The Self-Attention module in the Motion-Motion Interaction block is responsible for the processing of one of the interacting agents' motion features. Fig.~\ref{fig:fig_appendix_self_attn} is an illustration of this. The motion features are first modulated by an Adaptive Layer Norm (AdaLN) block, which injects timestep information by scaling the features with mean and variance determined by the timestep. Then, projection layers calculate the query, key, and value tensors separately, which are used to calculate the attention output using the attention mechanism~\citep{vaswani2017attention}. Finally, the attention output is projected with a linear output projection layer to give the final output. 

\subsubsection{Cross-Attention}
The Cross-Attention module in the Motion-Motion Interaction block is responsible for modeling the inter-agent interaction. Fig.~\ref{fig:fig_appendix_cross_attn} provides an illustration. In the Cross-Attention module, the motion features of agent A and agent B (vice versa) are used to calculate features using separate AdaLN blocks for the timestep information. Then, the motion features of agent A are passed through the query projection layer to give the query features. The motion features of agent B are passed through the key and value projection layer to calculate the key and value features. The query, key, and value features undergo an attention mechanism to obtain the attention output feature, which is subsequently projected by an linear output layer to form the final output.

\section{\namesyn Implementation Details}
\subsection{Two Person Prompt Synthesis}
We use a prompt template for synthesizing diverse and plausible two-person interaction descriptions from an LLM~\citep{liu2024deepseek} based on coarse-grained themes and fine-grained tags, along with real examples from the InterHuman~\citep{liang2024intergen} dataset for styling reference. The complete template is provided below in Fig.~\ref{fig:two_person_prompt_template}.

\begin{figure}[ht]
\centering

\begin{tcolorbox}[promptstyle]
You write compact, vivid descriptions of **two-person interactions**. 
\\\\
Each output sentence MUST:

• mention exactly two unnamed people (“one person… the other person…”),  

• focus on body / arms / legs (ignore faces / fingers / appearance),  

• be <=25 words,  

• clearly match the given *Theme* and *Tags*,  

• be entirely different from the examples.
\\\\

Theme: \textbf{\{theme\}}

Tags : \textbf{\{tags\}}

Reference examples ({k}):

\textbf{\{example1\}}\\
\textbf{\{example2\}}\\
\dots\\
\textbf{\{examplek\}}
\\\\
Now craft \textbf{\{m\}} brand-new descriptions.  
Return **only** a JSON array of strings.
\end{tcolorbox}

\caption{Prompt template for generating two-person interaction descriptions.}
\label{fig:two_person_prompt_template}
\end{figure}

\subsection{Single Person Prompt Synthesis}
After obtaining two-person descriptions, we use a separate prompt template to synthesize pairs of single-person descriptions that are self-contained, coherent, and consistent with the given two-person descriptions. The LLM infers the corresponding single-person motion according to the provided two-person interaction information, while reasoning the plausible single-person motion if the two-person prompt does not provide complete information. The complete prompt template is given in Fig.~\ref{fig:single_person_prompt_template}. Examples of two-person prompts and corresponding single-person prompts are given in Fig.~\ref{fig:prompt_examples}.

\begin{figure}[ht]
\centering

\begin{tcolorbox}[promptstyle]
Given the following description of a two-person interaction:
\\\\
\textbf{\{two-person text\}}
\\\\
Independently describe the motion of each person involved, using only information
implied by the full interaction. Do not mention or refer to the other person in
either description. Focus only on body, arms, and legs — ignore facial
expressions, fingers, or appearance.
\\\\
Use "the person" to refer to each. Assume shared context (e.g., dancing,
greeting, arguing), but isolate each description.
\\\\
Output JSON in this exact format:

\quad     \{\{"1": \{\{"person1": "\{description1\}", "person2": "\{description2\}"\}\}\}\}
\\\\
Each description must be one sentence, <=15 words, specific, and motion-focused
with relevant context.
\end{tcolorbox}

\caption{Prompt template for generating single-person interaction descriptions.}
\label{fig:single_person_prompt_template}
\end{figure}

\begin{figure}[ht]
\centering

\begin{tcolorbox}[promptstyle]
\textbf{Example 1.} \\
\textbf{two-person text}: One person leans back, arms outstretched, while the other steps forward, pressing their chest lightly against the first's, hands resting on their hips.\\
\textbf{single-person text A}: The person leans back with arms outstretched. \\
\textbf{single-person text B}: The person steps forward, chest pressed lightly, hands on hips.
\\\\
\textbf{Example 2.} \\
\textbf{two-person text}: One person claps twice, and the other responds by jumping in place, their legs kicking out wildly with excitement.\\
\textbf{single-person text A}: The person raises both arms and brings hands together sharply twice. \\
\textbf{single-person text B}: The person leaps upward, legs swinging outward vigorously.
\\\\
\textbf{Example 3.} \\
\textbf{two-person text}: One person lunges with a punch, the other person blocks with crossed arms and counters with a swift kick to the thigh.
\\
\textbf{single-person text A}: The person steps forward, extending one arm sharply in a punching motion. \\
\textbf{single-person text B}: The person raises both arms to cross in front, then swings one leg outward quickly.
\\\\
\textbf{Example 4.} \\
\textbf{two-person text}: one person steps forward aggressively, arms raised, while the other person backs away, hands outstretched to resist the advancing confrontation.
\\
\textbf{single-person text A}: The person steps forward aggressively with arms raised. \\
\textbf{single-person text B}: The person backs away with hands outstretched to resist.
\\\\
\textbf{Example 5.} \\
\textbf{two-person text}: One person stumbles backward from alcohol, and the other person swiftly wraps an arm around their waist to steady them.
\\
\textbf{single-person text A}: The person stumbles backward, legs unsteady from alcohol. \\
\textbf{single-person text B}: The person moves an arm quickly to wrap around a waist. 
\\\\
\textbf{Example 6.} \\
\textbf{two-person text}: One person shoves the other's shoulder, causing them to stagger, then crosses their arms in defiance as the other retreats.
\\
\textbf{single-person text A}: The person extends their arm sharply, then pulls it back and crosses both arms tightly. \\
\textbf{single-person text B}: The person stumbles backward from a sudden force, then turns away while stepping back.

\end{tcolorbox}

\caption{Examples of generated and two-person interaction descriptions with corresponding single-person descriptions.}
\label{fig:prompt_examples}
\end{figure}

\subsection{Single Person Motion Generation}
We use MoMask~\citep{guo2024momask} to generate the single-person motion conditions for the subsequent reaction generation. In addition, we trained a length estimator on the InterHuman~\citep{liang2024intergen} text-motion pairs to estimate the correct length of the corresponding motion given a two-person motion prompt, and used the predicted length by the estimator to guide the single-person motion generation.

For each two-person prompt, we use the LLM~\citep{liu2024deepseek} to provide two prompts for the two interactants and generate two single-person motions, one with each prompt.

\subsection{Two Person Motion Composition}
We trained a reaction generation network that uses a given sequence of joints as condition $\mathbf{x}_{cond}\in \mathbb{R}^{T \times 22 \times 3}$ to generate the reaction $\mathbf{x}\in \mathbb{R}^{T \times 262}$, consisting of the full interaction in the complete InterGen~\citep{liang2024intergen} joint representation. 

The network is trained on the InterHuman~\citep{liang2024intergen} training dataset with one person's joints not noised and all other terms noised to simulate the reaction generation tasks. At test time, the condition joints are provided at each time step of denoising and after the final denoising step. 

\subsection{Motion Filtering}
Before filtering, we first use the trained neural motion evaluator to project all generated motions to the $512$-dimensional motion latent space, to compare with the latents of a $500$-sample held-out motion dataset. The motion filtering step consists of a $k$-nearest neighbors filtering with a maximum $20$ nearest neighbors for each sample in the held-out set. We also calculate the distances between each of the nearest neighbors with the held-out motion sample to make sure the distance is in the predefined annulus $d_{min} < r < d_{max}$. Finally, we use the neural motion evaluator to filter out all the remaining motions with text-motion cosine similarity less than $0.58$, an empirically set threshold.

\section{Motion Embedding Space Visualizations}

\begin{figure*}[t]
\centering
\makebox[\textwidth][c]{
  \begin{overpic}[width=\linewidth]{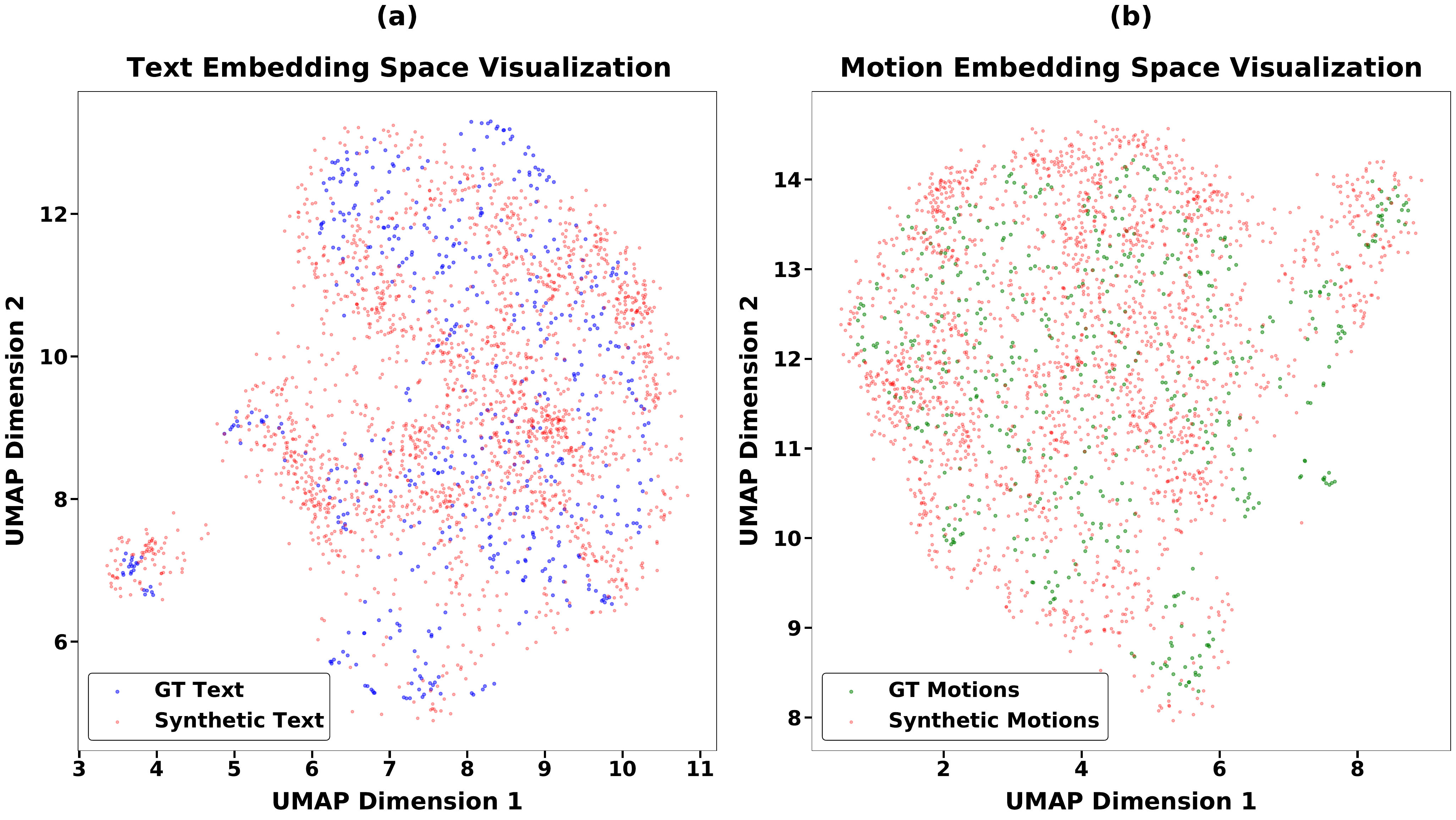}
  \end{overpic}
}
\caption{UMAP~\citep{mcinnes2018umap} visualizations of evaluator and CLIP~\citep{radford2021learning} embeddings of (a) text and (b) two-person motions from the InterHuman~\citep{liang2024intergen} held-out subset and filtered synthesized dataset. }
\label{fig:latent_space_visualization}
\end{figure*}

Fig.~\ref{fig:latent_space_visualization} demonstrates a dimensionality-reduced visualization of the text and two-person motion embeddings of the InterHuman~\citep{liang2024intergen} held-out dataset, extracted from CLIP~\citep{radford2021learning} and trained motion evaluator. We utilize UMAP~\citep{mcinnes2018umap} (Uniform Manifold Approximation and Projection), a popular dimension reduction technique that preserves the local and global structure of high-dimensional data in a low-dimensional space. As shown in the figure, both the generated text (Fig.~\ref{fig:latent_space_visualization} (a)) and motion (Fig.~\ref{fig:latent_space_visualization} (b)) descriptions from our pipeline have good coverage in most high-density areas of the held-out dataset, while covering many underrepresented areas that lacks held-out data samples, highlighting our pipeline's capability in enhancing data diversity. 

\section{User Study Details}
\begin{figure*}[htbp]
\centering
\makebox[\textwidth][c]{
  \begin{overpic}[width=0.95\linewidth]{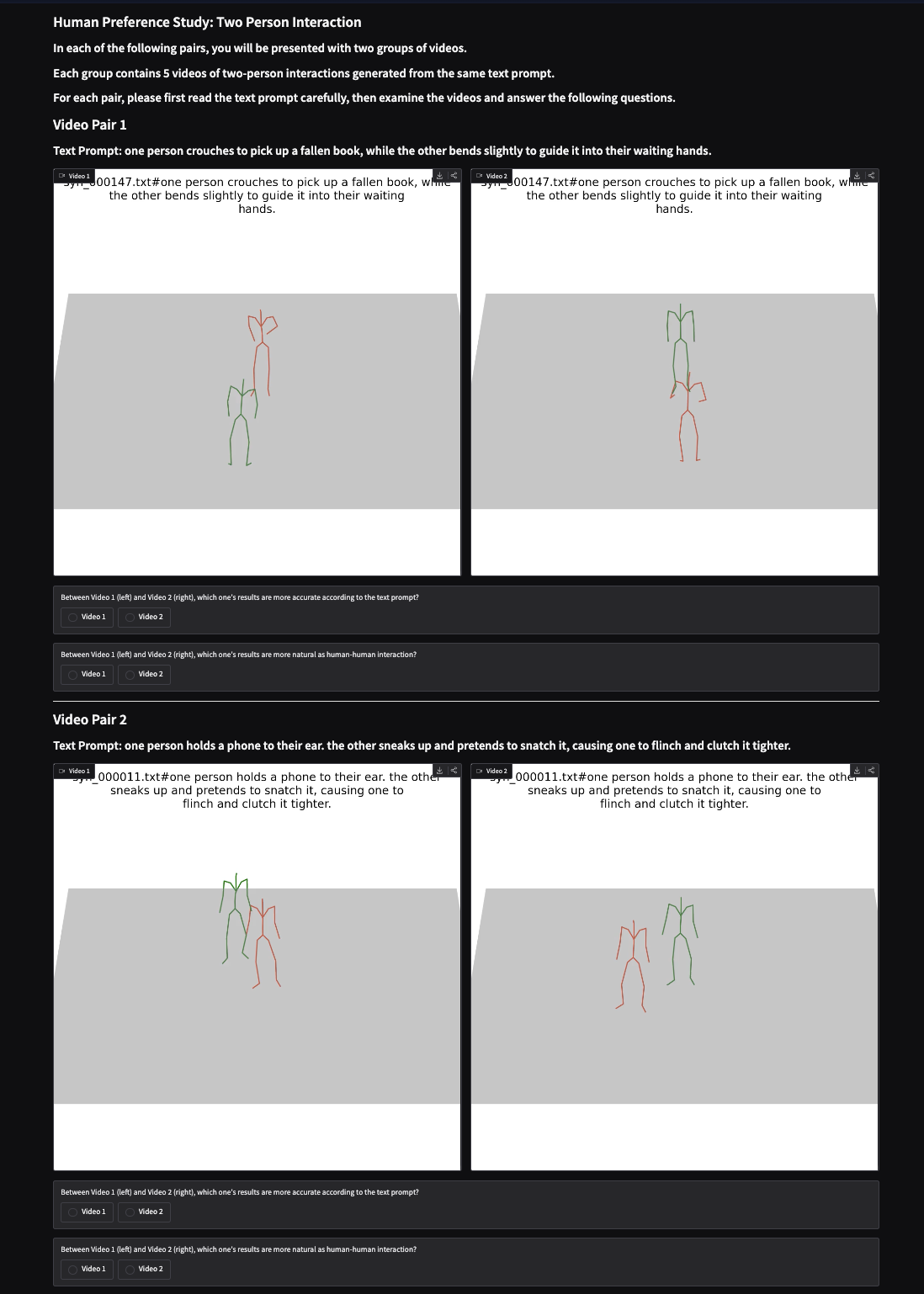}
  \end{overpic}
}
\caption{A screenshot of the user study}
\label{fig:fig_user_study}
\end{figure*}

We conducted a user study to evaluate our \namegen model with and without fine-tuning. Human evaluators are asked to choose between 10 pairs of two-person interaction videos and determine the one out of each pair that is more faithful to the text prompt or more natural as an interaction. Fig.~\ref{fig:fig_user_study} is an illustration of how the user study is conducted.

\section{Societal Impacts}
Our work on \name introduces a scalable framework for high-fidelity and diverse text-to-two-person interaction generation. While the technology has the potential to significantly benefit domains such as animation, virtual reality, assistive robotics, and embodied AI, it also raises several ethical and societal considerations.

\paragraph{Positive Impacts.} The proposed method can facilitate content creation in media, education, and human-computer interaction by automating the generation of complex, realistic human interactions. It may lower the barrier for creating high-quality motion data, particularly for under-resourced languages or motion types, and help simulate social interactions for training embodied agents or improving accessibility tools for individuals with disabilities.

\paragraph{Risks and Limitations.} As with many generative models, there is a potential risk of misuse, such as generating deceptive or misleading content (e.g., synthetic surveillance or manipulated footage). Although our method focuses solely on body motion and excludes facial expressions or identity features, generated motion could still be used out of context or embedded in misleading visual narratives. Additionally, there is a risk of dataset bias being amplified if the single-person priors or LLM-generated text reflect culturally specific or stereotyped behaviors. We recommend future users apply careful evaluation and transparency practices when deploying this technology.

\paragraph{Mitigations.} Our dataset curation and filtering process emphasizes diversity and alignment with real-world motion distributions to reduce representation biases. Furthermore, our model does not generate personally identifiable information or faces, and we encourage its use only in applications that respect human dignity, consent, and privacy.

\end{document}